\documentclass{article}
\usepackage{ijcai22}

\usepackage{times}
\usepackage{soul}
\usepackage{url}
\usepackage[hidelinks]{hyperref}
\usepackage[utf8]{inputenc}
\usepackage[small]{caption}
\usepackage{graphicx}
\usepackage{amsmath}
\usepackage{amsthm}
\usepackage{booktabs}
\usepackage{algorithm}
\usepackage{algorithmic}
\usepackage{enumitem}
\usepackage{amsfonts}
\usepackage{mathtools}
\usepackage{multirow}
\usepackage{booktabs}
\DeclarePairedDelimiter\floor{\lfloor}{\rfloor}

\title{Black-box Node Injection Attack for Graph Neural Networks}
\author {
    Mingxuan Ju$^1$ \and
    Yujie Fan$^2$ \and
    Yanfang Ye$^1$\thanks{Corresponding Author.} \and
    Liang Zhao$^3$
\affiliations
$^1$ University of Notre Dame, Notre Dame, IN 46556 \\
$^2$ Case Western Reserve University, Cleveland, OH 44106 \\
$^3$ Emory University, Atlanta, GA 30322 
\emails
\{mju2, yye7\}@nd.edu,
yxf370@case.edu,
liang.zhao@emory.edu
}

\begin{document}

\maketitle

\begin{abstract}
Graph Neural Networks (GNNs) have drawn significant attentions over the years and been broadly applied to vital fields that require high security standard such as product recommendation and traffic forecasting. Under such scenarios, exploiting GNN's vulnerabilities and further downgrade its classification performance become highly incentive for adversaries. Previous attackers mainly focus on structural perturbations of existing graphs. Although they deliver promising results, the actual implementation needs capability of manipulating the graph connectivity, which is impractical in some circumstances. In this work, we study the possibility of injecting nodes to evade the victim GNN model, and unlike previous related works with white-box setting, we significantly restrict the amount of accessible knowledge and explore the black-box setting. Specifically, we model the node injection attack as a Markov decision process and propose GA2C, a graph reinforcement learning framework in the fashion of advantage actor critic, to generate realistic features for injected nodes and seamlessly merge them into the original graph following the same topology characteristics. Through our extensive experiments on multiple acknowledged benchmark datasets, we demonstrate the superior performance of our proposed GA2C over existing state-of-the-art methods. The data and source code are publicly accessible at: \url{https://github.com/jumxglhf/GA2C}.
\end{abstract}

\section{Introduction}

Graph neural networks (GNNs), a class of deep learning methods designed to perform inference on graph data, have achieved outstanding performance in various real-world applications, such as text classification \cite{yao2019graph,peng2018large}, recommendation system \cite{ying2018graph,fan2019graph}, and traffic forecasting \cite{cui2019traffic,yu2017spatio}. The success of GNNs relies on their powerful capability of integrating the graph structure and node features simultaneously for node representation learning. Specifically, the majority of popular GNNs \cite{klicpera2019predict,kipf2016semi,hamilton2017inductive,velivckovic2017graph} follow a neural message-passing scheme to learn node embeddings via iteratively aggregating and propagating neighbor information. Along with their great success, the robustness of GNNs has also attracted increasing attentions in recent years. It is proved that the way that GNNs exchange information between nodes is vulnerable to adversarial attacks. Particularly, generating unnoticeable perturbations on graph structure or node features are able to degrade the learned node representations, and consequently deteriorate the performance of GNNs in downstream tasks. Existing research efforts on graph adversarial attack mainly concentrate on graph structure perturbation via modifying edges \cite{dai2018adversarial,ma2019attacking,wang2019attacking,xu2019topology,zugner2018adversarial,zugner2018adversarial}. Despite their promising performance, the attack strategies used in these works have narrow applications where the adversaries are required to have capability of manipulating the graph connectivity. Besides graph structural perturbations, node injection attacks on graphs are also gaining attentions. This type of attack focuses on a more practical scenario where the attackers don't need to have high authority in modifying existing graph structure. Considering the spam detection in social network as an example, attackers don't have ability to protect the spam account via adding or removing the already formed friendships among existing users. But they are able to generate new accounts with new profiles, and establish new links with existing users to fulfill the attack purpose. Apparently, such attack is less expensive and more feasible compared with the attacks from graph structure perturbation. Nevertheless, node injection attack is a challenging task and the adversaries are expected to consider: (i) \textit{how to generate imperceptible discrete features for the injected fake node?} and (ii) \textit{how to establish links between an injected node and the existing nodes in the original graph?} Current works \cite{wang2020scalable,tao2021single,sun2019node} achieve this type of attack in a white-box or grey-box setting where the attackers have full or partially limited knowledge about the training inputs and labels, the model parameters and gradient information. Nonetheless, such white-box and grey-box attacks are usually not practical for real-world adversaries.

In this work, we consider a more challenging and practical scenario, i.e., black-box node injection attack, where the attackers are only allowed to access the adjacency and attribute matrices. To the best of our knowledge, this is the first attempt that formally investigates the node injection attack under the black-box setting. Such attack could be simply regarded as a search process with exponentially large search space, which can be possibly solved with brute force search. However, from the perspective of practicality, a machine learning aided framework that approximates the optimal solution of the search process should be utilized. To tackle these aforementioned challenges, based on advanced actor critic (A2C), we propose a reinforcement learning-based attack framework, namely GA2C, that requires only queries from the victim model. Specifically, we formulate the node injection attack as a Markov Decision Process (MDP), where the attack is decoupled into node generation and edge wiring. During the node generation phase, imperceptible discrete features are attributed to the adversarial node through a special case of Gumbel-Softmax. And during the edge wiring phase, edges between the injected node and the remaining graph are sampled from a learnable conditional probability distribution. The key contributions of this paper are summarized as:
\begin{itemize}[leftmargin=*]
    \item This is the first work that formally studies node injection attack for GNNs under the black-box setting, where only very restrained knowledge is accessible. 
    \item We carefully formulate the node injection attack as a MDP, and aided with designated effective reward function, we propose a graph reinforcement learning framework, namely GA2C, to approximate the optimal solution under extremely limited black-box setting.
    \item Through comprehensive comparison experiments on multiple acknowledged benchmark datasets, we demonstrate GA2C's superior attack efficiency under different attack budgets,by comparison with state-of-the-art attack models. 
\end{itemize}

\begin{figure*}[!h]
\centering
\includegraphics[width=1\linewidth]{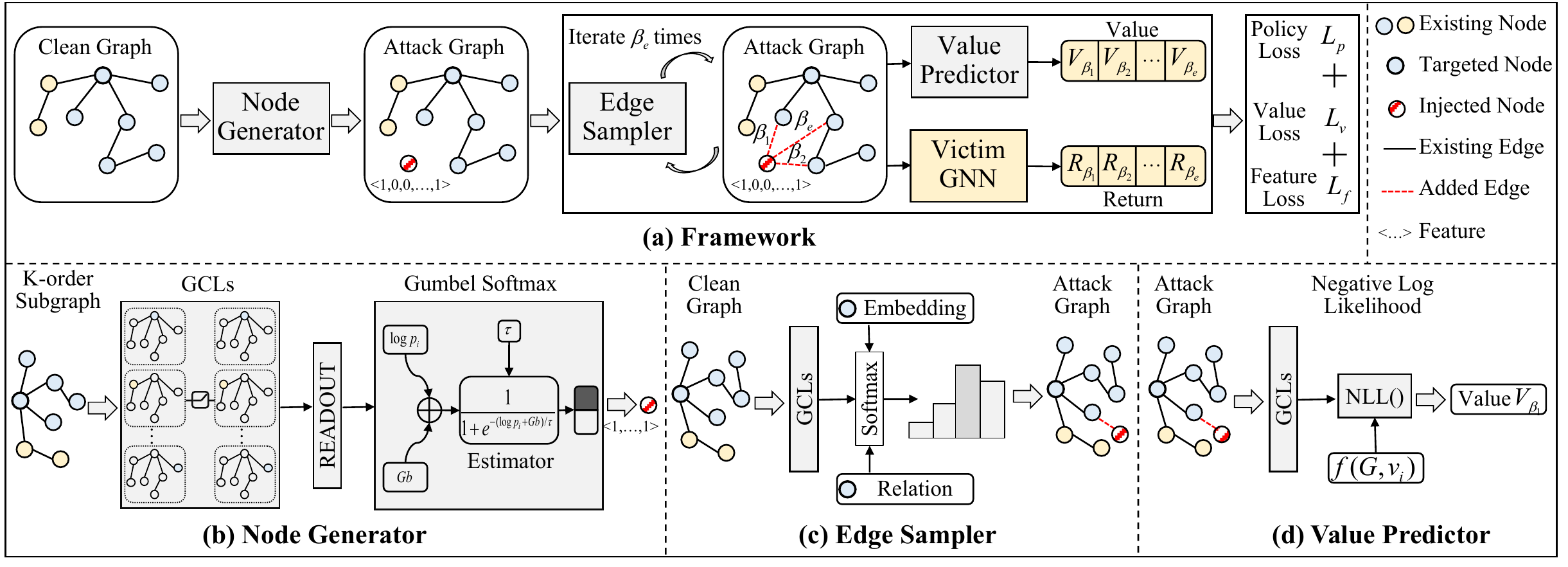}
\caption{System overview of GA2C.}
\vspace{-0.4cm}
\label{fig:system}
\end{figure*}

\section{Preliminary}

Let $G = (V,E)$ denote a graph, where $V$ is the set of $|V| = N$ nodes and $E \subseteq V \times V$ is the set of $|E|$ edges between nodes. Adjacency matrix is denoted as $\textbf{A} \subseteq \{0,1\}^{N \times N}$, where $a_{ij}$ at $i$-th row and $j$-th column equals to 1 if there exists an edge between nodes $v_i$ and $v_j$ or equals to 0 otherwise. We further denote the node feature matrix as $\mathbf{X} \in \mathbb{R}^{N \times F}$ where node $v_i$ is given with a feature vector $\mathbf{x}_i \in \mathbb{R}^F$ and $F$ is the dimension size. $\mathbf{Y} \subseteq \{0,1\}^{N \times C}$ denotes the label matrix of a graph, where $C$ is the number of total classes. For $M$ labeled nodes ($0 < M \ll N$) with label $\mathbf{Y}^L$ and $N-M$ unlabeled nodes with missing label $\mathbf{Y}^U$, the objective of GNNs for node classification is to predict $\mathbf{Y}^U$ given $\mathbf{Y}^L$, $\mathbf{A}$ and $\mathbf{X}$. 

\subsection{Graph Neural Network}

GNNs generalize neural networks into graph-structured data \cite{scarselli2008graph,kipf2016semi,klicpera2019predict,velivckovic2017graph}. The key operation is graph convolution where information is routed between nodes with some deterministic rules (e.g., adjacency matrices and Laplacian matrices). For example, the graph convolution layer (GCL) of GCN \cite{kipf2016semi} is formulated as: $\mathbf{H}^{(l+1)} = f_{GCL}^{(l)}(\hat{\mathbf{A}}, \mathbf{H}^{(l)},\mathbf{W}^{(l)})= \sigma(\hat{\mathbf{A}} \mathbf{H}^{(l)} \mathbf{W}^{(l)})$,
where $\hat{\mathbf{A}}$ denotes the normalized adjacency matrix with self-loop, $\sigma(.)$ denotes the non-linearity function, and $\mathbf{W}^{(l)}$ and $\mathbf{H}^{(l)}$is the learnable parameters and node representations at $l^{th}$ layer respectively. Normally, at $K$-th layer, with the last dimension of $\mathbf{W}^{(K)}$ and $\sigma(.)$ set to $C$ and softmax respectively, the loss for node $v_i$ is formulated as:
\begin{equation}
	\mathcal{L}(v_i, G, \mathbf{y}_i) = CE\big( f_{GCL}^{(K)}(\hat{\mathbf{A}}, \mathbf{H}^{(K)},\mathbf{W}^{(K)})[i], \mathbf{y}_i\big),
\end{equation} 
where $[\cdot]$ is the indexing operation and $CE(\cdot,\cdot)$ refers to cross entropy loss function.

\subsection{Graph Adversarial Attack}
For a trained GNN model $f(\cdot,\cdot): V \times G \rightarrow \mathbf{Y}$, the attacker $g(\cdot,\cdot): G \times f \rightarrow G$ asks to modify the graph $G=(V,E)$ into $G'=(V', E')$ such that:
\begin{equation}
	\begin{split}
		\max_{G'} \;\;\;\;& \mathbb{I}(f(V^U, G') \neq \mathbf{Y}^U) \\
		s.t. \;\;\;\; & G' = g(G,f) \text{ and } \mathcal{I}(G, G') = 1.
	\end{split}
\label{eq:constraint}
\end{equation}
Here $V^U$ can be the testing set or nodes of interest, $\mathbb{I}(\cdot)$ is an indicator function that returns the number of true conditions, and $\mathcal{I}(\cdot, \cdot):G\times G \rightarrow \{0,1\}$ is an indicator function, which returns one if two graphs are equivalent under the classification semantics. Besides, there are two graph components that can be attacked as targets. First target is edge modification in $G$, also known as structural attack, which changes the entries in $\mathbf{A}$. Whereas the second target tampers the nodes via adding, modifying, or deleting nodes in $G$, resulting in not only entry-level but also dimensional changes to both $\mathbf{A}$ and $\mathbf{X}$. In this work, we focus on black-box node injection evasion attack, a special case of the second target, where three attack budgets need to be considered: the number of adversarial nodes injected per attacking one node, denoted as $\beta_n$, the degree of adversarial node $\beta_e$, and lastly the number of features assigned to each adversarial node $\beta_f$. Hence, the indicator function $\mathcal{I}(\cdot, \cdot)$ in Eq. (\ref{eq:constraint}) becomes $\mathcal{I}(G, G') = \mathbb{I}(|V'| - |V| \leq \beta_n) \cdot \prod_{i=N+1}^{N+\beta_n}\mathbb{I}(\sum_{j=1}^{N} a_{ij}' \leq \beta_e) \cdot  \prod_{i=N+1}^{N+\beta_n}\mathbb{I}(\sum_{j=1}^{F} x_{ij}' \leq \beta_f).$

\section{Methodology}
Given the clean graph $G = (V,E)$, the attacker $g$ injects adversarial evasion node set $V^{\mathcal{A}}$ with its adversarial features $\mathbf{X}^{\mathcal{A}}$ into the clean node set $V$. After injecting $V^{\mathcal{A}}$, attacker $g$ creates adversarial edges $E^{\mathcal{A}} \subseteq V^{\mathcal{A}} \times V \cup V^{\mathcal{A}} \times V^{\mathcal{A}}$ to evade the detection of GNN $f$ for nodes $V^U$. $G' = (V', E')$ is the attacked graph in which $V' = V \cup V^{\mathcal{A}}$, $E' = E \cup E^{\mathcal{A}}$, and $\mathbf{X}' = \mathbf{X} \oplus \mathbf{X}^{\mathcal{A}}$, where $\oplus$ is the append operation.

Injecting node involves generating discrete graph data, such as discrete adjacency matrix or feature matrix, that gradient-based approaches handle poorly under many circumstances. This phenomenon could be further aggravated by the black-box setting where gradient information from surrogate model is not accurate. Moreover, generating node and assigning edges are naturally sequential and reinforcement learning fits for such Markov Decision Process (MDP). Hence, to perform such optimization task in Eq. (\ref{eq:constraint}), we propose to explore deep reinforcement learning. Specifically, we utilize an on-policy A2C reinforcement learning framework, adapted from \cite{mnih2016asynchronous}, instead of off-policy algorithms such as deep Q-learning. Since A2C circumvents the needs to calculate the expected future value for every possible feature combinations, which are exponentially large (e.g., Q learning computes $F \choose \beta_f$ Q values, which exceeds $10^{41}$ possibilities with common selection of $F=1000$ and $\beta_f = 20$; whereas on-policy A2C requires only one step). 

The overview of GA2C is shown in Figure \ref{fig:system}. Given graph $G$ and target node $v_i$, the node generator $g_n$ attributes the adversarial node according to topological information of the $v_i$'s sub-graph. After that, edge sampler $g_e$ is forwarded for $\beta_e$ consecutive times to connect the introduced adversarial nodes to the original graphs. Previous two processes recursively iterate until $\beta_n$ adversarial nodes have been injected or the label of $v_i$ has been successfully changed. A detailed definition of our proposed MDP is defined as follows:

\paragraph{State.} $s_t \in S$ contains the intermediate attacked graph $G'_t = (V'_t, E'_t)$ as well as generated features $\mathbf{X}^{\mathcal{A}}_t$ at timestamp $t$. To effectively interpret $s_t$, edge sampler $g_e$ attends to whole $G'_t$ and could connect $v_i$ to any node that could bring optimal perturbations. However, for node generator $g_n$, it firstly extracts the $K$-order sub-graph $G'_t(v_i)$ of $v_i$, where $K$ is a hyper-parameter for the number of stacked GCLs in $g_n$. We limited our scope on neighbors within $K$ hops since normally the victim GNN $f$ has shallow receptive field.

\paragraph{Action.} Node injection attack can be decoupled into two components: attributing the adversarial node and connecting this node to the remaining of attacked graph. We model this process as a MDP where GA2C starts with node attributing and then conducts edge wiring from injected node to the remaining graph for $\beta_e$ consecutive times. The MDP terminates at any time if the attacker $g$ successfully evades the detection from $f$ or a total number of $\beta_n$ nodes are injected. Formally, at time $t$, the node generation action is denoted as $a^{(n)}_t$, and the node wiring action is denoted as $a^{(e)}_t$. The trajectory of our proposed MDP is ($s_0$, $a^{(n)}_0$, $s_1$, $a^{(e)}_1$, $r_1$, $s_2$, $a^{(e)}_2$, $r_2$, $s_2$, $\dots$,$a^{(n)}_t$, $s_t$, $\dots$, $s_T$), where $s_T$ refers to the terminal state, and $r_t$ is the reward for action $a_t$. 

\paragraph{Reward.} In our designed setup, generating an isolated node does not deliver a reward value as it would not bring any perturbation to $f$; its impact is reflected later when links are wired to the original graph. Hence, a reward value is only assigned to edge wiring actions. During the intermediate phase of attacking node $v_i$ at timestamp $t$, the reward for edge wiring action $a^{(e)}_t$ is calculated as:
\begin{equation}
\begin{split}
  r(v_i, a^{(e)}_t, G'_t) &= \mathcal{L}\big(v_i, G'_{t+1}, f(v_i, G)\big)\\
  &- \mathcal{L}\big(v_i, G'_t, f(v_i, G)\big) \;\; \text{if}\; s_{t+1} \neq s_T,
\end{split}
  \label{eq:reward1}
\end{equation}
where $G'_{t+1}$ is the resulted graph of applying $a^{(e)}_t$ to $G'_t$. Furthermore, to encourage GA2C not only minimizing the decision boundaries as enforced by rewards calculated by increase in classification loss function but also evading the detection of $f$, we give extra rewards if the prediction of $v_i$ is flipped at the end of one attacking episode (i.e., $\mathbb{I}\big(f(G'_T, v_i) \neq f(v_i, G)\big)$).

\subsection{Node Injection Evasion Attack via Actor Critic}

A2C includes two modules: an on-policy model that outputs optimal action given current state, and a value predictor that derives the actual accumulated reward from current timestamp to the end of this MDP. To adapt A2C to the node injection evasion attack, we design three components: node generator $g_n$, edge sampler $g_e$ and value predictor $g_v$. 

\paragraph{Adversarial Node Generator}

Given a target node $v_i$, and its K-order sub-graph $G'_t(v_i)$, node generator $g_n$ aims to attribute the injected adversarial node $v_a$ with a feature vector $\mathbf{x}_a$, while considering the imperceptibility of $\mathbf{x}_a$. Specifically, the generated $\mathbf{x}_a$ from $g_n$ should follow the same characteristic conventions as $\mathbf{X}$. We shouldn't expect a continuous $\mathbf{x}_a$ when all other feature vectors are discrete, vice and versa. In this paper, we study the special and far less researched discrete feature generation, since binary feature vectors are commonly seen in the real world. Moreover, $\mathbf{x}_a$ should not diverge too much from nodes features in $G'_t(v_i)$. To tackle the aforementioned challenges, $g_n$ is equipped with $K$ stacked GCLs, conducts message propagation on $G'_t(v_i)$, summarizes $G'_t(v_i)$ by a readout function \cite{xu2018powerful}, and with Gumbel-Softmax \cite{jang2016categorical}, generates $\mathbf{x}_a$ tailoring the vulnerability of $v_i$ as well as the imperceptibility. Formally, the $K$-th convolution layer of $g_n$ can be described as: $ \mathbf{H}^{(K+1)}_n = f_{GCL}^{(n,K)}(\hat{\mathbf{A}}(v_i), \mathbf{H}^{(K)}_n,\mathbf{W}^{(K)}_n)$,
where $ \mathbf{H}^{(0)}_n = \{\mathbf{x}_i | v_i \in G'_t(v_i)\}$, $\hat{\mathbf{A}}(v_i)$ refers the normalized adjacency matrix of $G'_t(v_i)$ and $\mathbf{W}^{(K)}_n$ is the parameter of $g_n$'s $K$-th convolution layer. To gather the topological information of the whole $G'_t(v_i)$ and as well as the unique characteristics tailored by $v_i$, we formulate the feature distribution $\mathbf{z}_n$ as:
\begin{equation*}
    \mathbf{z}_n = \sigma\Big(\big(READOUT(\mathbf{H}^{(K+1)}_n)||\mathbf{H}^{(K+1)}_n(v_i)\big) \cdot \mathbf{W}_n^f\Big),
\end{equation*}
where $READOUT(.)$ refers to graph pooling readout function such as column-wise max or summation \cite{xu2018powerful}, $\mathbf{H}^{(K+1)}_n(v_i)$ denotes $v_i$'s node embedding after propagation, and $\mathbf{W}_n^f \in \mathbb{R}^{2d \times F}$ is the learnable parameters that combines local topological information and target node's characteristics. Then, we utilize $\mathbf{z}_n$ as the logits of a relaxed Bernoulli probability distribution, a binary special case of the Gumbel-Softmax reparameterization trick that is soft and differentiable \cite{jang2016categorical}. Utilizing the relaxed sample, we apply a straight-through gradient estimator \cite{bengio2013estimating}, that rounds the relaxed sample in the forward phase. In the backward propagation, actual gradients are directly passed to relaxed samples instead of previously rounded values, making $g_n$ trainable. Formally, $\mathbf{x}_a$ can be generated by:
\vspace{-0.3cm}
\begin{equation}
\begin{split}
        \mathbf{x}_a[i] &= \floor*{\frac{1}{1+e^{-(\log p_i + Gb)/\tau}} + \frac{1}{2}},\\
        \text{s.t.} \;\;\;& p_i = \alpha_n \cdot \mathbf{z}_n[i] + (1-\alpha_n) \cdot \mathbf{h}^{(0)}_n[i]
\end{split}
\label{eq:gumbel}
\end{equation}
where $[\cdot]$ is the indexing operation, $Gb \sim Gumbel(0,1)$ is a Gumbel random variable, $\tau$ is the temperature of Gumbel-Softmax distribution, $\alpha_n$ is the hyper-parameter adjusting the influence of $G'_t(v_i)$'s feature distribution, and $\mathbf{h}^{(0)}_n$ is the mean average of $\mathbf{H}^{(0)}_n$. Besides, to maintain the imperceptibility of $\mathbf{x}_a$, we propose a feature loss function: $\mathcal{L}_f(\mathbf{x}_a) = \big((\sum_{i=0}^{F} \mathbf{x}_a[i]) - \beta_f\big)^2$.

\paragraph{Adversarial Edge Sampler}
While attacking node $v_i$, given generated node features $\mathbf{x}_a$ from $g_n$, the adversarial edge sampler $g_e$ aims at connecting $v_a$ to the rest of graph $G'_t$. Similar to $g_n$, $g_e$ is also equipped with a $K$ stacked GCLs, except that $g_e$ takes the whole $G'_t$ as input, allowing it to wire $v_a$ to any other node in $G'_t$,  formulated as: $\mathbf{H}^{(K+1)}_e = f_{GCL}^{(e,K)}(\hat{\mathbf{A}}', \mathbf{H}^{(K)}_e,\mathbf{W}^{(K)}_e)$, where $\mathbf{H}^{(0)}_e = \mathbf{X}'$, $\hat{\mathbf{A}}'$ refers the normalized adjacency matrix of $G'_t$ and $\mathbf{W}^{(K)}_e$ is the parameter of $g_e$'s $K$-th convolution layer. Then, we concatenate $\mathbf{x}_a$ with each row of $\mathbf{H}^{(K+1)}_e$ and receives $\mathbf{Z}_e \in \mathbb{R}^{|V'| \times (d+F)}$. The probability vector of remaining nodes connecting to $v_a$ is calculated as:
\begin{equation}
    \mathbf{p}_e = \text{softmax}(\mathbf{Z}_e \cdot \mathbf{W}_e + \mathbf{A}'(v_i)),
    \label{eq:edge_prob}
\end{equation}
where $\mathbf{W}_e \in \mathbb{R}^{(d+F)}$ is the learable parameters and $\mathbf{A}'(v_i)$ denotes $v_i$'s row in the adjacency matrix of $G'_t$. We add $\mathbf{A}'(v_i)$ to the probability logits because in order for the adversarial perturbation being perceived by $f$, the introduced edge must enable $\mathbf{v}_a$ to lie in the receptive field of $f$. Manually increasing the probabilities of $v_a$ connecting to the first-order neighbors of $v_i$ speeds up the convergence time. Then, we sample an edge from an one-hot categorical distribution parameterized by $\mathbf{p}_e$, merge the sampled edge into $G'_t$ and get $G'_{t+1}$. For the next edge sampling operation, $G'_{t+1}$ is fed into $g_e$, and this process iterates until $v_i$ is successfully evaded or the number of wired edges reaches $\beta_e$. 

\paragraph{Value Predictor}

Along with policy learners $g_n$ and $g_e$ we have proposed, the value predictor $g_v$ is the other component of A2C that aims at predicting the expected accumulated rewards at the end of MDP. Given the dedicated reward functions we define in Eq. (\ref{eq:reward1}), $g_v$ should predict the final accumulated loss score of targeted node based on current $G'_t$. We formulate this process as a regression task, where $g_v$ predicts the negative log likelihood between the class log probabilities in current graph $G'_t$ and $f(G, v_i)$. Firstly, a GNN model with $K$ stacked layer is utilized to capture the node topological information, similar to $g_e$, as: $\mathbf{H}^{(K+1)}_v = f_{GCL}^{(v,K)}(\hat{\mathbf{A}}', \mathbf{H}^{(K)}_v,\mathbf{W}^{(K)}_v)$, where $\mathbf{W}^{(K)}_v$ is the parameter of $g_v$'s K-th convolution layer. Then, we extract $v_i$'s node embedding and concatenate it with $f$'s output to predict the value score, formulated as:
\begin{equation*}
 g_v(v_i, G'_t) = NLL\Big(\big(\mathbf{H}^{(K+1)}_v(v_i)||f(v_i, G'_t)\big)\cdot \mathbf{W}_v, f(v_i, G)\Big),
\end{equation*}
where $NLL(\cdot, \cdot)$ is the negative log likelihood function, and $\mathbf{W}_v \in \mathbb{R}^{(d+C) \times C}$ is the learnable parameter.
\begin{table*}[h]
\centering
\begin{tabular}{c|ccc|ccc|ccc}
    \toprule
    \toprule
    \multirow{2}{*}{Method}    &  & Citeseer  &  &  & Cora &  &  &  Pubmed \\
    \cmidrule(r){2-4} 
    \cmidrule(r){5-7}
    \cmidrule(r){8-10}
         & $\beta_n = 1$ & $\beta_n = 2$ & $\beta_n = 3$ & $\beta_n = 1$ & $\beta_n = 2$ & $\beta_n = 3$ &  $\beta_n = 1$ & $\beta_n = 2$ & $\beta_n = 3$ \\
    \midrule
    Random        & 70.2 & 69.9 & 69.1 & 81.1 & 80.8 & 80.2 & 79.0 & 78.4 & 77.2\\
    Node+Rand     & 68.2 & 66.7 & 63.1 & 78.2 & 77.4 & 75.8 & 74.4 & 73.6 & 70.9\\
    Rand+Edge     & 68.8 & 67.2 & 64.4 & 78.8 & 77.9 & 76.3 & 75.3 & 74.2 & 71.5\\
    Rand+Nettack  & 68.7 & 67.2 & 65.3 & 79.2 & 78.1 & 77.6 & 76.0 & 75.3 & 73.4\\
    Node+Nettack  & 65.3 & 60.9 & 55.2 & 75.5 & 72.3 & 69.6 & 71.2 & 64.9 & 59.9\\
    NIPA          & 69.1 & 67.3 & 65.9 & 80.9 & 79.2 & 77.8 & 77.4 & 75.3 & 74.0\\
    Node+NIPA     & 68.2 & 63.6 & 61.1 & 77.4 & 74.8 & 71.4 & 75.7 & 71.2 & 68.2\\
    G-NIA         & 65.7 & \underline{59.1} & 55.3 & 75.2 & 73.1 & 69.3 & \underline{70.2} & \underline{64.2} & \underline{57.2}\\
    AFGSM         & \underline{65.2} & 59.4 & \underline{54.8} & \underline{74.6} & \underline{71.2} & \underline{66.3} & 71.5 & 66.2 & 59.3\\
    \midrule
    GA2C (ours)   & \textbf{62.6} & \textbf{57.6} & \textbf{51.8} &2 \textbf{70.8} & \textbf{66.2} & \textbf{57.1} & \textbf{61.3} & \textbf{53.3} & \textbf{52.8}\\
    \bottomrule
    \bottomrule
  \end{tabular}
  \caption{Classification by GCN under different attackers (\%, lower is better. \textbf{Accuracy} in bold indicates the best and \underline{accuracy} in underline is the second best).}
  \label{tab:exp_overall}
  \vspace{-0.2cm}
 \end{table*}
\subsection{Training Algorithm}
To train GA2C = $\{g_n, g_e, g_v\}$, we explore the experience replay technique with memory buffer $\mathcal{M}$. Intuitively, we simulate the selection process to generate training data, store the experience in the memory buffer during the forward runs of training phase. An instance in $\mathcal{M}$ is in the format of triplet $(G'_t, a_t, R_t)$ with return $R_t = \sum_{j=t}^{j=T} r(v_i, a_j, G'_j) \cdot \gamma^{(j-t)}$, where $\gamma$ refers to the discount factor. During the back-propagation phase, three losses are involved: policy loss $\mathcal{L}_p$, value loss $\mathcal{L}_v$ and feature loss $\mathcal{L}_f$. Given a triplet $(G'_t, a_t, R_t) \in \mathcal{M}$, policy loss is calculated as:
\begin{equation*}
     \mathcal{L}_p(G'_t, a_t, R_t) = - \log\big(p(a_t|G'_t)\big) \cdot \big(R_t - g_v(v_i, G'_t)\big),
\end{equation*}
where $p(a_t|G'_t)$ denotes the probability of conducting action $a_t$ under the graph $G'_t$. In $\mathcal{L}_p$, the second term $(R_t - g_v(v_i, G'_t))$ is also known as the advantage score \cite{mnih2016asynchronous}, which depicts how much better of selecting action $a_t$ over other actions. $\mathcal{L}_p$ enforces GA2C to deliver better actions with higher probabilities. On the other hand, value loss $\mathcal{L}_v$ enforces the value predictor $g_v$ to correctly deliver the actual accumulated reward, calculated as: 
\begin{equation*}
     \mathcal{L}_v(G'_t, R_t) = |g_v(v_i, G'_t) - R_t|.
\end{equation*}
The final loss for GA2C is formulated as:
\begin{equation*}
\mathcal{L} = \sum_{\mathcal{M}} \mathcal{L}_v(G'_t, R_t) + \mathcal{L}_p(G'_t, a_t, R_t) + \sum_{\mathbf{x}_a \in \mathbf{X}_t^{\mathcal{A}}} \mathcal{L}_f(\mathbf{x}_a)
\end{equation*}

\section{Experiment}
In this section, we aim at answering the following four research questions: (\textbf{RQ1}) Can our proposed GA2C effectively evade target nodes given a trained GNN? (\textbf{RQ2}) How do the key component in GA2C impact the performance? (\textbf{RQ3}) What is the attack performance of GA2C under different budgets? (\textbf{RQ4}) How does GA2C conduct node injection attack?

\paragraph{Datasets.} We conduct experiments on three acknowledged benchmark datasets, namely Cora, Citeseer and Pubmed \cite{sen2008collective}. We follow the community convention \cite{kipf2016semi} and explore the public splits (e.g., fixed 20 nodes per class for training, 500 for validation and 1,000 for testing). We take GCN as the victim model and attackers aim at downgrading its testing accuracy. The dataset statistics and performance of GCN on clean graphs are shown in Table \ref{dataset}. 
 \begin{table}[h]
    \centering
    \begin{tabular}{c|ccccc}
    \toprule
    Dataset & Node & Edge & Class & Feat. & Acc.\\
    \midrule
    Cora     & 2,708 & 5,429 & 7 & 1,433 & 81.5\\
    Citeseer & 3,327 & 4,732 & 6 & 3,703 & 70.3\\
    Pubmed   & 19,717 & 44,338 & 3 & 500 & 79.0\\
    \bottomrule
    \end{tabular}
    \caption{Benchmark dataset statistics.}
    \label{dataset}
\end{table}
\vspace{-0.2cm}
\paragraph{Baselines.} Since the node injection attack is an emerging and far less researched area, only few methods focus on this topic such as NIPA \cite{sun2019node}, G-NIA \cite{tao2021single}, and AFGSM \cite{wang2020scalable}. To demonstrate the effectiveness of GA2C, besides existing methods, we also compare GA2C with adaptions of the state-of-the-art edge perturbation methods. In sum, our baselines include: three variant of random attack (i.e., Random, Node+Rand., and Rand.+Edge), NIPA \cite{sun2019node} and its variant (i.e., Node+NIPA), two variants of Nettack \cite{zugner2018adversarial} (i.e., Rand.+Nettack and Node+Nettack), G-NIA \cite{tao2021single} and AFGSM \cite{wang2020scalable}. The description of each baseline is given in the appendi. To compare GA2C with other white-box (i.e., Nettack, AFGSM and G-NIA) or grey-box (i.e., NIPA) approaches that require the gradient information from victim model, we adopt a common solution of training a surrogate model.

\subsection{Performance Comparison}

The performances of GA2C and baselines are reported in Table \ref{tab:exp_overall}. We observe that all baselines can cause a performance downgrade to victim model, including the most basic random-based methods. By comparing models equipped with random features with those utilizing dedicated adversarial features (e.g., Rand+Nettack vs. Node+Nettack), we can clearly observe the adversarial features greatly imperil the performance of victim model, and most baselines can cause considerable downgrade with reasonably crafted features. The performance of G-NIA and AFGSM is jeopardized significantly due to the black-box setting, comparing to their originally reported performance. To answer \textbf{RQ1}: under the black-box setting, GA2C outperforms all baselines and by a significant margin on Pubmed. The performance of weak baselines can benefit from the adversarial features generated by GA2C, as demonstrated by Node+Rand, Node+NIPA and Node+Nettack. By comparing Node+Nettack with GA2C, we can also observe a higher performance for GA2C, indicating the outstanding edge wiring capability of edge sampler. 

\subsection{Ablation Study}
Besides node generator $g_n$ and edge sampler $g_e$, whose performance is already demonstrated in Table \ref{tab:exp_overall}, to answer \textbf{RQ2}, we examine the efficacy of other components integrated in GA2C. To investigate the contribution of Gumbel-Softmax in Eq. (\ref{eq:gumbel}), we replace it with a vanilla Top-K operation, denoted as $\text{GA2C}_{\text{w/o GS}}$. Nevertheless, in order to justify the design of proposed reward function, we also design a reward based on accuracy, calculated as the increase of accuracy between two intermediate graphs, denoted as $\text{GA2C}_{\text{acc}}$. Furthermore, in Eq. (\ref{eq:edge_prob}), we manually add the information of target node for fast convergence, and we also experiment on a variant, denoted as $\text{GA2C}_{\text{w/o pre}}$, to test if GA2C can converge under no prior information. From Table \ref{tab:ablation}, we observe that $\text{GA2C}_{\text{w/o GS}}$ without designated feature generation equipped with Gumbel-Softmax suffers significantly, due to the fact that during each back-propagation, only a portion of the learnable matrix $\mathbf{W}_n^f$ can be updated. For $\text{GA2C}_{\text{acc}}$, the attacking performance is severely downgraded, because no immediate rewards are given for actions that could not directly flip the label of target node. Moreover, although $\text{GA2C}_{\text{w/o pre}}$ achieves similar performance as regular GA2C, we experience a noticeable more training time on this variant.
\begin{table}[h]
\centering
\vspace{-0.25cm}
 \begin{tabular}{c|ccc}
    \toprule
    Variant & Citeseer & Cora & Pubmed \\
    \midrule
    GA2C                           & 51.8 & 57.1 & 52.8\\
    $\text{GA2C}_{\text{w/o GS}}$  & 55.1 & 69.5 & 58.3\\
    $\text{GA2C}_{\text{acc}}$     & 56.4 & 71.3 & 59.8\\
    $\text{GA2C}_{\text{w/o pre}}$ & 51.6 & 57.2 & 53.1\\
    \bottomrule
  \end{tabular}
  \caption{Ablation studies on GA2C (\%).  $\beta_n = 3$}
  \label{tab:ablation}
 \end{table}
 \vspace{-0.25cm}
\subsection{Budget Analysis}
\begin{figure}[h]
\centering
\includegraphics[width=1\linewidth]{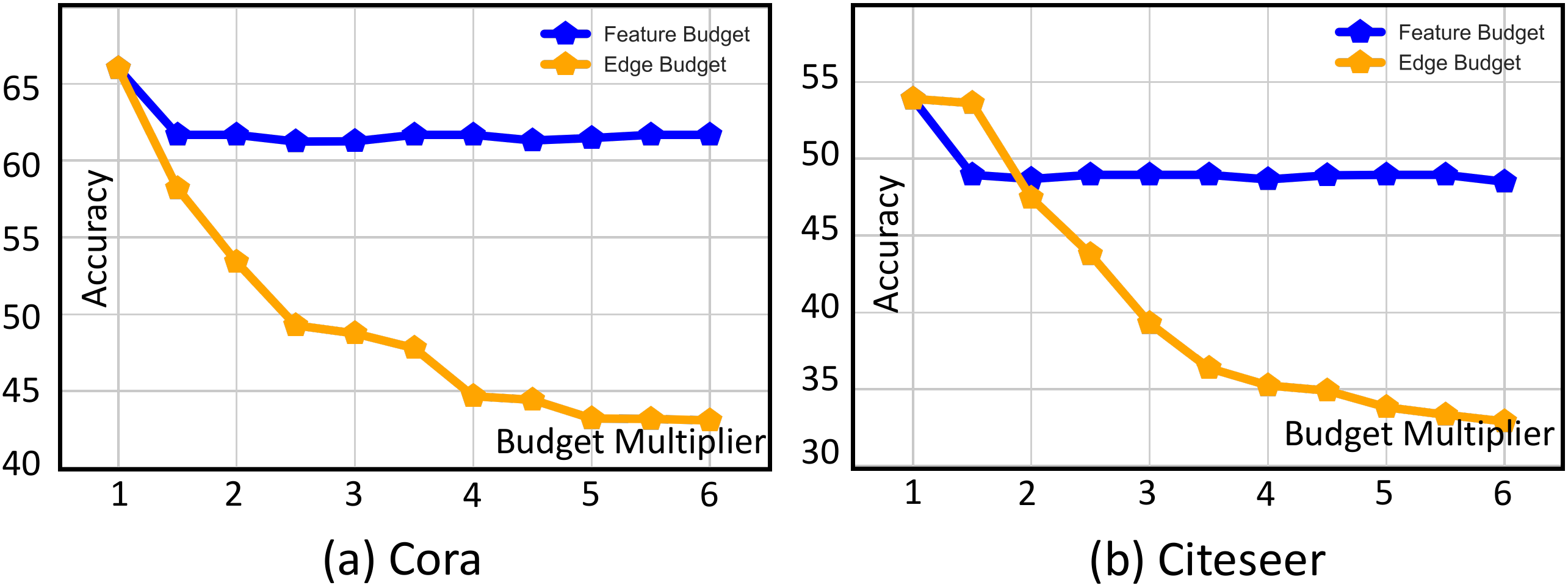}
\caption{GA2C with different budget (\%).}
\label{fig:budget}
\end{figure}

Besides the number of injected nodes $\beta_n$ evaluated in Table \ref{tab:exp_overall}, in this section, we also conduct experiment on the number of allowed edges per each node $\beta_e$, and the number of allowed features per node $\beta_f$ as shown in Figure \ref{fig:budget}. From these results, to answer \textbf{RQ3}, the most fruitful budget is $\beta_n$, and with 3 injected nodes per target, the performance of GCN on all datasets is downgraded to the range of 50\%s. The second most impactful budget is $\beta_e$, and the performance of GCN all falls down below 50\% with 2 times edge budget on Citeseer, and with 2.5 times edge budget on Cora. The least impactful budget is $\beta_f$, from which we observe that on all dataset, its impact saturates with 1.5 times budget. 
\subsection{Case Study}
\begin{figure}[h]
\centering
\includegraphics[width=1\linewidth]{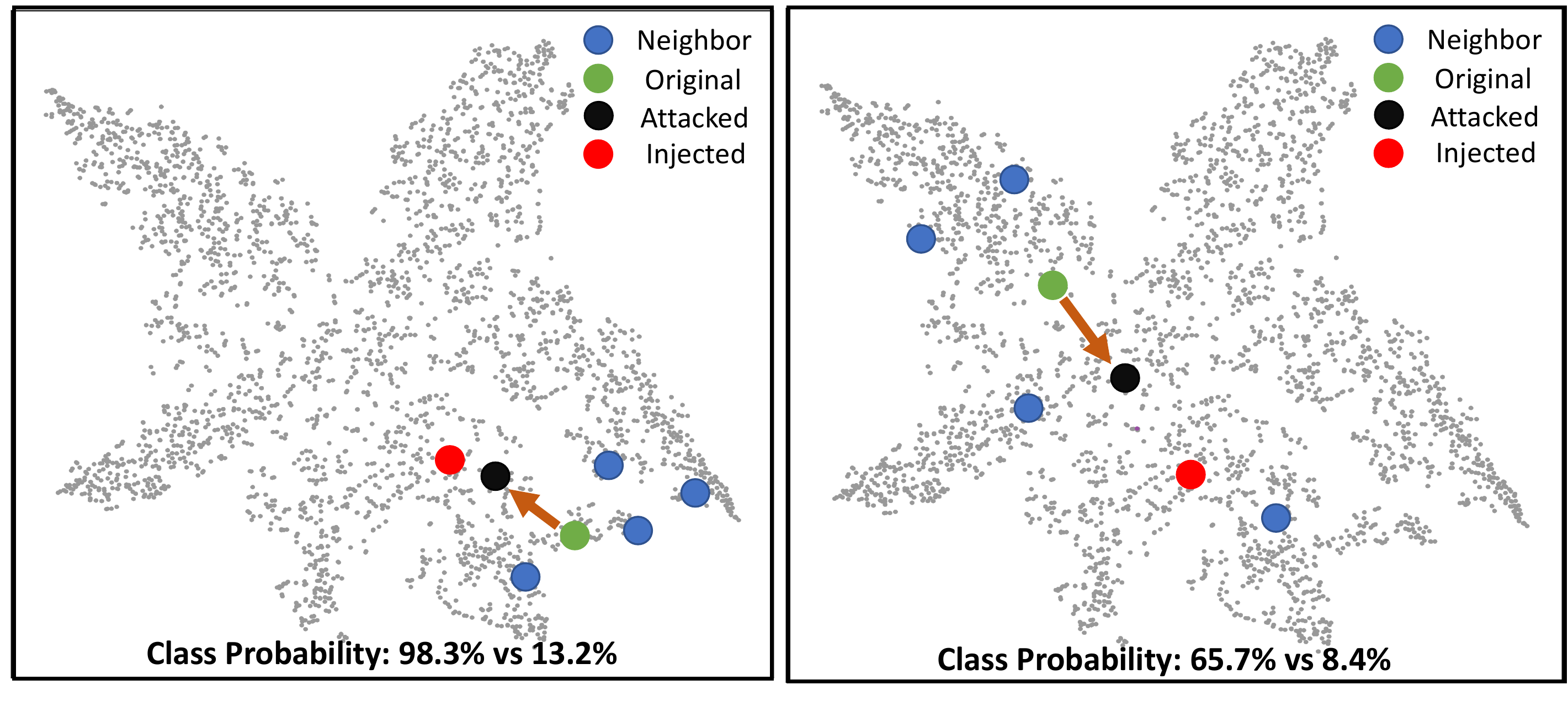}
\caption{Visualization of node injection attack by GA2C.}
\label{fig:tsne}
\end{figure}

To further investigate how GA2C conducts node injection attack, we visualize two successful attacks on Citeseer. As shown in Figure \ref{fig:tsne}, we visualize the attack process by plotting the hidden embedding of involved nodes, extracted from the victim model, before and after the attack via T-SNE \cite{van2008visualizing}. In this figure, blue points are target node's original neighbors in the clean graph, green point is the target node before the attack, black point refers to the attacked target node, and red point refers to the injected adversarial node. To answer \textbf{RQ4}, as shown in these two cases, the injected node could effectively perturb the embedding of target node, relocate the target node to a relatively intertwined position, and flip the prediction of target node.

\section{Related Work}
GNNs have been proved to be sensitive to adversarial attacks \cite{dai2018adversarial,ma2019attacking,wang2019attacking,xu2019topology,zugner2018adversarial,sun2019node,tao2021single,wang2020scalable}. Most of these attacks focus on perturbations on existing knowledge, such as topological structure \cite{xu2019topology,wang2019attacking,ma2019attacking}, node attributes \cite{zugner2018adversarial}, and labels \cite{sun2019node}. For example, CE-PGD \cite{xu2019topology} and Nettack \cite{zugner2018adversarial} exploit gradient to modify the graph structure and attributes. However, in the real world, modifying existing edges or node attributes is not practical, due to limited access to node of interests. Node injection attack aims at a more realistic scenario, which adds adversarial nodes in to the existing graph. State-of-the-art attackers \cite{sun2019node,tao2021single,wang2020scalable} explore a white- \cite{tao2021single,wang2020scalable} or grey-box \cite{sun2019node} scenario, where model gradient and training label are accessible. Specifically, AFGSM \cite{wang2020scalable} utilize a fast gradient sign method to inject node and G-NIA \cite{tao2021single} explores a neural network to generalize the attacking process. To further leverage the practicality of node injection attack, we study it under the black-box setting, a more restricted scenario where only adjacency and feature matrices are available. 

\section{Conclusion}
In this paper, we study the problem of black-box node injection evasion attack on graph-structured data and propose GA2C. With little knowledge of target model and training data, we formulate such attack as a MDP and solve it through our designed graph reinforcement learning framework based on A2C. During the node injection phase, a node generator is utilized to generate adversarial yet realistic node features, followed by a edge sampler that wires generated node to the remaining graph. Through extensive comparison experiments with state-of-the-art baselines, we demonstrate that GA2C outperforms all of them on three widely acknowledged datasets. We also discuss the influence of different attack budgets on GA2C and show that it could achieve satisfactory results with limited budgets. Our future work includes designing generalizable reinforcement learning model for graph adversarial attack.

\newpage

\bibliographystyle{named}
\bibliography{ijcai22}

\begin{thebibliography}{}

\bibitem[\protect\citeauthoryear{Bengio \bgroup \em et al.\egroup
  }{2013}]{bengio2013estimating}
Yoshua Bengio, Nicholas L{\'e}onard, and Aaron Courville.
\newblock Estimating or propagating gradients through stochastic neurons for
  conditional computation.
\newblock {\em arXiv preprint arXiv:1308.3432}, 2013.

\bibitem[\protect\citeauthoryear{Cui \bgroup \em et al.\egroup
  }{2019}]{cui2019traffic}
Zhiyong Cui, Kristian Henrickson, Ruimin Ke, and Yinhai Wang.
\newblock Traffic graph convolutional recurrent neural network: A deep learning
  framework for network-scale traffic learning and forecasting.
\newblock {\em IEEE Transactions on Intelligent Transportation Systems},
  21(11):4883--4894, 2019.

\bibitem[\protect\citeauthoryear{Dai \bgroup \em et al.\egroup
  }{2018}]{dai2018adversarial}
Hanjun Dai, Hui Li, Tian Tian, Xin Huang, Lin Wang, Jun Zhu, and Le~Song.
\newblock Adversarial attack on graph structured data.
\newblock In {\em International conference on machine learning}, pages
  1115--1124. PMLR, 2018.

\bibitem[\protect\citeauthoryear{Fan \bgroup \em et al.\egroup
  }{2019}]{fan2019graph}
Wenqi Fan, Yao Ma, Qing Li, Yuan He, Eric Zhao, Jiliang Tang, and Dawei Yin.
\newblock Graph neural networks for social recommendation.
\newblock In {\em The World Wide Web Conference}, pages 417--426, 2019.

\bibitem[\protect\citeauthoryear{Hamilton \bgroup \em et al.\egroup
  }{2017}]{hamilton2017inductive}
William~L Hamilton, Rex Ying, and Jure Leskovec.
\newblock Inductive representation learning on large graphs.
\newblock In {\em Proceedings of the 31st International Conference on Neural
  Information Processing Systems}, pages 1025--1035, 2017.

\bibitem[\protect\citeauthoryear{Jang \bgroup \em et al.\egroup
  }{2016}]{jang2016categorical}
Eric Jang, Shixiang Gu, and Ben Poole.
\newblock Categorical reparameterization with gumbel-softmax.
\newblock {\em arXiv preprint arXiv:1611.01144}, 2016.

\bibitem[\protect\citeauthoryear{Kipf and Welling}{2016}]{kipf2016semi}
Thomas~N Kipf and Max Welling.
\newblock Semi-supervised classification with graph convolutional networks.
\newblock {\em arXiv preprint arXiv:1609.02907}, 2016.

\bibitem[\protect\citeauthoryear{Klicpera \bgroup \em et al.\egroup
  }{2019}]{klicpera2019predict}
Johannes Klicpera, Aleksandar Bojchevski, and Stephan G{\"u}nnemann.
\newblock Predict then propagate: Graph neural networks meet personalized
  pagerank.
\newblock 2019.

\bibitem[\protect\citeauthoryear{Ma \bgroup \em et al.\egroup
  }{2019}]{ma2019attacking}
Yao Ma, Suhang Wang, Tyler Derr, Lingfei Wu, and Jiliang Tang.
\newblock Attacking graph convolutional networks via rewiring.
\newblock {\em arXiv preprint arXiv:1906.03750}, 2019.

\bibitem[\protect\citeauthoryear{Mnih \bgroup \em et al.\egroup
  }{2016}]{mnih2016asynchronous}
Volodymyr Mnih, Adria~Puigdomenech Badia, Mehdi Mirza, Alex Graves, Timothy
  Lillicrap, Tim Harley, David Silver, and Koray Kavukcuoglu.
\newblock Asynchronous methods for deep reinforcement learning.
\newblock In {\em International conference on machine learning}, pages
  1928--1937. PMLR, 2016.

\bibitem[\protect\citeauthoryear{Peng \bgroup \em et al.\egroup
  }{2018}]{peng2018large}
Hao Peng, Jianxin Li, Yu~He, Yaopeng Liu, Mengjiao Bao, Lihong Wang, Yangqiu
  Song, and Qiang Yang.
\newblock Large-scale hierarchical text classification with recursively
  regularized deep graph-cnn.
\newblock In {\em Proceedings of the 2018 world wide web conference}, pages
  1063--1072, 2018.

\bibitem[\protect\citeauthoryear{Scarselli \bgroup \em et al.\egroup
  }{2008}]{scarselli2008graph}
Franco Scarselli, Marco Gori, Ah~Chung Tsoi, Markus Hagenbuchner, and Gabriele
  Monfardini.
\newblock The graph neural network model.
\newblock {\em IEEE transactions on neural networks}, 20(1):61--80, 2008.

\bibitem[\protect\citeauthoryear{Sen \bgroup \em et al.\egroup
  }{2008}]{sen2008collective}
Prithviraj Sen, Galileo Namata, Mustafa Bilgic, Lise Getoor, Brian Galligher,
  and Tina Eliassi-Rad.
\newblock Collective classification in network data.
\newblock {\em AI magazine}, 29(3):93--93, 2008.

\bibitem[\protect\citeauthoryear{Sun \bgroup \em et al.\egroup
  }{2019}]{sun2019node}
Yiwei Sun, Suhang Wang, Xianfeng Tang, Tsung-Yu Hsieh, and Vasant Honavar.
\newblock Node injection attacks on graphs via reinforcement learning.
\newblock {\em arXiv preprint arXiv:1909.06543}, 2019.

\bibitem[\protect\citeauthoryear{Tao \bgroup \em et al.\egroup
  }{2021}]{tao2021single}
Shuchang Tao, Qi~Cao, Huawei Shen, Junjie Huang, Yunfan Wu, and Xueqi Cheng.
\newblock Single node injection attack against graph neural networks.
\newblock {\em arXiv preprint arXiv:2108.13049}, 2021.

\bibitem[\protect\citeauthoryear{Van~der Maaten and
  Hinton}{2008}]{van2008visualizing}
Laurens Van~der Maaten and Geoffrey Hinton.
\newblock Visualizing data using t-sne.
\newblock {\em Journal of machine learning research}, 9(11), 2008.

\bibitem[\protect\citeauthoryear{Veli{\v{c}}kovi{\'c} \bgroup \em et al.\egroup
  }{2017}]{velivckovic2017graph}
Petar Veli{\v{c}}kovi{\'c}, Guillem Cucurull, Arantxa Casanova, Adriana Romero,
  Pietro Lio, and Yoshua Bengio.
\newblock Graph attention networks.
\newblock {\em arXiv preprint arXiv:1710.10903}, 2017.

\bibitem[\protect\citeauthoryear{Wang and Gong}{2019}]{wang2019attacking}
Binghui Wang and Neil~Zhenqiang Gong.
\newblock Attacking graph-based classification via manipulating the graph
  structure.
\newblock In {\em Proceedings of the 2019 ACM SIGSAC Conference on Computer and
  Communications Security}, pages 2023--2040, 2019.

\bibitem[\protect\citeauthoryear{Wang \bgroup \em et al.\egroup
  }{2020}]{wang2020scalable}
Jihong Wang, Minnan Luo, Fnu Suya, Jundong Li, Zijiang Yang, and Qinghua Zheng.
\newblock Scalable attack on graph data by injecting vicious nodes.
\newblock {\em Data Mining and Knowledge Discovery}, 34(5):1363--1389, 2020.

\bibitem[\protect\citeauthoryear{Xu \bgroup \em et al.\egroup
  }{2018}]{xu2018powerful}
Keyulu Xu, Weihua Hu, Jure Leskovec, and Stefanie Jegelka.
\newblock How powerful are graph neural networks?
\newblock {\em arXiv preprint arXiv:1810.00826}, 2018.

\bibitem[\protect\citeauthoryear{Xu \bgroup \em et al.\egroup
  }{2019}]{xu2019topology}
Kaidi Xu, Hongge Chen, Sijia Liu, Pin-Yu Chen, Tsui-Wei Weng, Mingyi Hong, and
  Xue Lin.
\newblock Topology attack and defense for graph neural networks: An
  optimization perspective.
\newblock {\em arXiv preprint arXiv:1906.04214}, 2019.

\bibitem[\protect\citeauthoryear{Yao \bgroup \em et al.\egroup
  }{2019}]{yao2019graph}
Liang Yao, Chengsheng Mao, and Yuan Luo.
\newblock Graph convolutional networks for text classification.
\newblock In {\em Proceedings of the AAAI conference on artificial
  intelligence}, volume~33, pages 7370--7377, 2019.

\bibitem[\protect\citeauthoryear{Ying \bgroup \em et al.\egroup
  }{2018}]{ying2018graph}
Rex Ying, Ruining He, Kaifeng Chen, Pong Eksombatchai, William~L Hamilton, and
  Jure Leskovec.
\newblock Graph convolutional neural networks for web-scale recommender
  systems.
\newblock In {\em Proceedings of the 24th ACM SIGKDD International Conference
  on Knowledge Discovery \& Data Mining}, pages 974--983, 2018.

\bibitem[\protect\citeauthoryear{Yu \bgroup \em et al.\egroup
  }{2017}]{yu2017spatio}
Bing Yu, Haoteng Yin, and Zhanxing Zhu.
\newblock Spatio-temporal graph convolutional networks: A deep learning
  framework for traffic forecasting.
\newblock {\em arXiv preprint arXiv:1709.04875}, 2017.

\bibitem[\protect\citeauthoryear{Z{\"u}gner \bgroup \em et al.\egroup
  }{2018}]{zugner2018adversarial}
Daniel Z{\"u}gner, Amir Akbarnejad, and Stephan G{\"u}nnemann.
\newblock Adversarial attacks on neural networks for graph data.
\newblock In {\em Proceedings of the 24th ACM SIGKDD International Conference
  on Knowledge Discovery \& Data Mining}, pages 2847--2856, 2018.

\end{thebibliography}

\makebox[0.5 \textwidth]{\textbf{\Large GA2C Supplementary Appendix}}

\section{Baseline Description}

\noindent In this section, we provide a detailed description of each baseline used in the experiment.
\begin{itemize}[leftmargin=*]
    \item \textbf{Random}: We design three random attack variants. The first variant, denoted as \textit{Random}, is implemented with both random node generation and edge wiring. For the second variant, denoted as \textit{Node+random}, we inject the nodes generated by GA2C, and randomly connect them to the target nodes or their corresponding first-order neighbors. The third variant utilizes the edge wiring module in GA2C with randomly generated nodes, denoted as \textit{Random+edge}.
    \item \textbf{Nettack}: Nettack \cite{zugner2018adversarial} is a white-box poisoning attack that firstly studies adversarial machine learning on graph-structured data. It perturbs features and edges of nodes in existing graphs that maximally downgrades the performance of victim model. To adapt Nettack into our scenario, we only allow edge modifications on the injected nodes and disregard the feature modification functionality. Two variant \textit{Random+Nettack} and \textit{Node+Nettack} are designed, similar to \textit{Node+random}. 
    \item \textbf{NIPA}: NIPA \cite{sun2019node} aims at poisoning attacks via node injection. Features of the injected nodes are sampled from a Gaussian distribution whose mean is the average feature value and variance is one. Edges from injected nodes to the remaining graph are inferred through Q-learning. We are concerned with the capability and practicality of such feature generation process, and besides regular NIPA, we also design a variant, \textit{Node+NIPA}, where the features of injected nodes are provided by GA2C. 
    \item \textbf{AFGSM}: AFGSM \cite{wang2020scalable} aims at white-box node injection evasion attack. It calculates the approximation of optimal solutions for attacking GNN. Within the attack budgets, it selects attributes and edges that could maximally perturb the loss of target nodes. 
    \item \textbf{G-NIA}: G-NIA \cite{tao2021single} also focuses at white-box node injection evasion attack. It utilizes the generalization ability of neural network and proposes a gradient-based generalizable attack strategy. Similar to AFGSM, it selects attributes and edges that could maximally perturb the loss of target nodes. 
\end{itemize}

As for the attack budget, besides the number of injected nodes $\beta_n$ displayed already, all other budgets are set to the average numbers of the original datasets (i.e., edges per injected node $\beta_e$ to the average degree, and the number of features to the average summation of original feature vectors). 

\section{Experimental Setup} 

\noindent The hyper-parameters for victim model GCN are the same as described in the original work \cite{kipf2016semi}. For the implementation of baselines, we explore DeepRobust library as well as public source code from the authors and use default settings. We set the hyper-parameters in GA2C as following: the number of GCLs $K$ to 2, the temperature parameter of Gumbel-Softmax to 1.0, hidden dimension $d$ to 64, the balancing term in Gumbel Softmax to 0.5, and the discount factor to 0.95. We train GA2C with batch size equals to 10, and utilize Adam optimizer with learning rate $10^{-3}$. Besides, we adopt an early stopping strategy, where GA2C stops training if the testing accuracy stops increasing for 100 continuous epochs and the best model is saved. All experiments are conducted 10 times and mean value is reported. The version of software we use is Python 3.8.8, PyTorch 1.9.0 and NumPy 1.20.1. All experiments run on a server with AMD Ryzen 3990X CPU, RTX A6000 GPU with 48GB RAM, and 128GB RAM.

\section{Detailed Training Procedure of GA2C}

\noindent In the section of methodology, we present the training procedure of GA2C and this section explains it with more detail, shown in Algorithm \ref{alg:algorithm}. The algorithm only demonstrates back propagation for one batch with size 1, and it could be easily extended to batches with larger size. In this algorithm, line 3 depicts the node generation process. Line 5-9 indicate the process of wiring one edge, while storing the value and intermediate reward. If during the edge wiring process, the label of target node is flipped, the attacking is terminated and we conduct parameter updating for GA2C, as shown by line 10-11 and line 15-17. We further investigate the time complexity of GA2C. The complexity of $g_n$, $g_e$, $g_v$ is upper-bounded by the graph convolution layer, which has complexity of $\mathcal{O}(|E|\cdot d)$. To injected $\beta_n$ nodes, each node is maximally capable of being wired to $\beta_e$ edges, and hence the complexity of GA2C becomes $\mathcal{O}(\beta_n \cdot \beta_e \cdot |E|\cdot d)$. Specifically, the node generator $g_n$ and value predictor $g_f$ take a K-hop reachable sub-graph as input, enabling the scalability of these two components invariant to the global graph. As for the edge sampler $g_v$, its scalability is the same as vanilla GCN. 
\begin{algorithm}[!h]
\caption{Training Procedure of GA2C}
\label{alg:algorithm}
\textbf{Input}: Clean graph $G$, victim GNN $f$, target node $v_i$, attack budgets $\{ \beta_n, \beta_e, \beta_f \}$\\
\textbf{Parameter}: $g_n$, $g_e$, $g_f$\\
\textbf{Output}: Updated model $g_n$, $g_e$, $g_f$
\begin{algorithmic}[1] 
\STATE  Memory buffer $\mathcal{M} = \{\}$, reward buffer $\mathcal{R} = \{\}$, value buffer  $\mathcal{V} = \{\}$, $G'\leftarrow G$
\FOR{$n=1$ to $\beta_n$}
\STATE $G' \leftarrow G' \cup g_n(v_i, G' | \beta_f)$
\FOR{$e=1$ to $\beta_e$}
\STATE $\mathcal{V} \leftarrow \mathcal{V} \cup g_v(v_i, G')$
\STATE action = $g_e(v_i, G')$
\STATE $\mathcal{R} \leftarrow \mathcal{R} \cup r(v_i, \text{action}, G')$
\STATE $\mathcal{M} \leftarrow \mathcal{M} \cup (\text{action}, G')$
\STATE $G' \leftarrow  action$
\IF{$f(v_i, G') \neq f(G, v_i)$}
\STATE Go to line 15
\ENDIF
\ENDFOR
\ENDFOR
\STATE Calculate $R_t$, and update $\mathcal{M}$
\STATE $g_n, g_e, g_f \leftarrow Backprop(\mathcal{L}_f,\mathcal{L}_v,\mathcal{L}_p)$
\STATE \textbf{return} $g_n$, $g_e$, $g_f$
\end{algorithmic}
\end{algorithm}

\subsection{Deeper Insights into the Loss Function}

\noindent To further understand of our designed loss, we explain the intuition behind every individual term in the final loss function. We start with the first and third term, two terms that are relatively easy to follow. The first term $\mathcal{L}_v$ is the value loss that enforces the value predictor $g_v$ to correctly predict the testing loss of target model after the attack. The third term $\mathcal{L}_f$ prevents $g_n$ from generating an excessive amount of features. The tricky part is the second term $\mathcal{L}_p$, the policy loss. It enforces GA2C to deliver fruitful actions with high probability. To give an intuitive example, let's assume that a terrible action is made with a very high probability. Such circumstance results in a low $-\log\big(p(a_t|G'_t)\big)$ and negative $\big(R_t - g_v(v_i, G'_t)\big)$, making $\mathcal{L}_p$ negative. However, to further decrease $\mathcal{L}_p$, $-\log\big(p(a_t|G'_t)\big)$ needs to increase, which in turn decreases $p(a_t|G'_t)$ and makes the probability of this terrible action low. The intertwinement between value predictor and action generators is similar to the mini-max game in generative adversarial networks, where the discriminator judges the outputs from the generator. Here, action generators learns from the value predictor such that actions with high accumulated future rewards are selected more frequently. 

\section{Limitation}

\noindent Reinforcement learning-based framework that models MDP could not utilize the parallelism computation and suffers from the efficiency downgrade brought by the sequential inference. Hence major limitation is support for parallel computation. 


\end{document}